\documentclass{vgtc}                          

\ifpdf
\pdfoutput=1\relax                   
\usepackage{graphicx}                
\DeclareGraphicsExtensions{.pdf,.png,.jpg,.jpeg} 
\else
\usepackage{graphicx}                
\DeclareGraphicsExtensions{.eps}     
\fi%

\graphicspath{{figures/}} 

\usepackage{microtype}                 
\PassOptionsToPackage{warn}{textcomp}  
\usepackage{textcomp}                  
\usepackage{mathptmx}                  
\usepackage{times}                     
\usepackage{cite}                      
\usepackage{tabu}                      
\usepackage{booktabs}                  

\onlineid{1012}

\vgtccategory{Research}

\vgtcinsertpkg

\usepackage{times}
\usepackage{epsfig}
\usepackage{graphicx}
\usepackage{amsmath}
\usepackage{amssymb}
\usepackage{textcomp}
\usepackage{subcaption}
\usepackage{bm}

\usepackage{float}

\usepackage{algorithm}
\usepackage{algpseudocode}

\renewcommand{\vec}[1]{\mathbf{#1}}
\newcommand{\mat}[1]{\mathbf{#1}}
\newcommand{\vecg}[1]{\boldsymbol{#1}} 

\DeclareMathOperator*{\argmin}{arg\,min}

\def\eg{e.g.}
\newcommand{\etal}{\textit{et al}. }

\usepackage{fancyhdr}

\begin{document}
	\thispagestyle{fancy}
	\fancyhead[C]{\textit{To appear in Adjunct Proceedings of the IEEE International Symposium for Mixed and Augmented Reality 2018}}
	\pagestyle{plain}

	\setlength{\belowdisplayskip}{3pt} \setlength{\belowdisplayshortskip}{3pt}
	\setlength{\abovedisplayskip}{3pt} \setlength{\abovedisplayshortskip}{3pt}
	
	\title{A Single-shot-per-pose Camera-Projector Calibration System For Imperfect Planar Targets}
	
	\author{Bingyao Huang \thanks{e-mail: bingyao.huang@temple.edu}\\
		\scriptsize	Temple Univeristy\\
		\and Samed Ozdemir \thanks{e-mail: ozdemi63@students.rowan.edu}\\
		\scriptsize	Rowan Univeristy\\
		\and Ying Tang \thanks{e-mail: tang@rowan.edu}\\
		\scriptsize	Rowan Univeristy\\
		\and Chunyuan Liao \thanks{e-mail: liaocy@hiscene.com}\\
		\scriptsize	HiScene Info.\ Technologies\\
		\and Haibin Ling \thanks{e-mail: hbling@temple.edu}\\
		\scriptsize	Temple Univeristy\\
	}
	
	\abstract{
		Existing camera-projector calibration methods typically warp feature points from a camera image to a projector image using estimated homographies, and often suffer from errors in camera parameters and noise due to imperfect planarity of the calibration target.
		In this paper we propose a simple yet robust solution that explicitly deals with these challenges. Following the structured light (SL) camera-project calibration framework, a carefully designed correspondence algorithm is built on top of the De Bruijn patterns. Such correspondence is then used for initial camera-projector calibration. Then, to gain more robustness against noises, especially those from an imperfect planar calibration board, a bundle adjustment algorithm is developed to jointly optimize the estimated camera and projector models.
		Aside from the robustness, our solution requires only one shot of SL pattern for each calibration board pose, which is much more convenient than multi-shot solutions in practice.
		Data validations are conducted on both synthetic and real datasets, and our method shows clear advantages over existing methods in all experiments.}
	
	\CCScatlist
	{
		\CCScatTwelve{Computing methodologies}{Camera calibration}{}{};
		\CCScatTwelve{Computing methodologies}{3D imaging}{}{};
		\CCScatTwelve{Computing methodologies}{Reconstruction}{}{}
	}
	\maketitle
	
	\thispagestyle{fancy}

	\section{Introduction}\label{sec:introduction}
	Camera-projector systems are popular in 3D surface reconstruction and projected augmented reality, where in most cases, structured light (SL) is applied due to its ease of use and accuracy. Compared with passive feature point based 3D reconstruction methods, such as stereo vision and structure from motion (SfM), SL is able to reconstruct a denser and more precise surface. Moreover, SL works for texture-less or repetitively textured objects.
	
\begin{figure}[!t]
		\centering
		\includegraphics[width=1.0\linewidth]{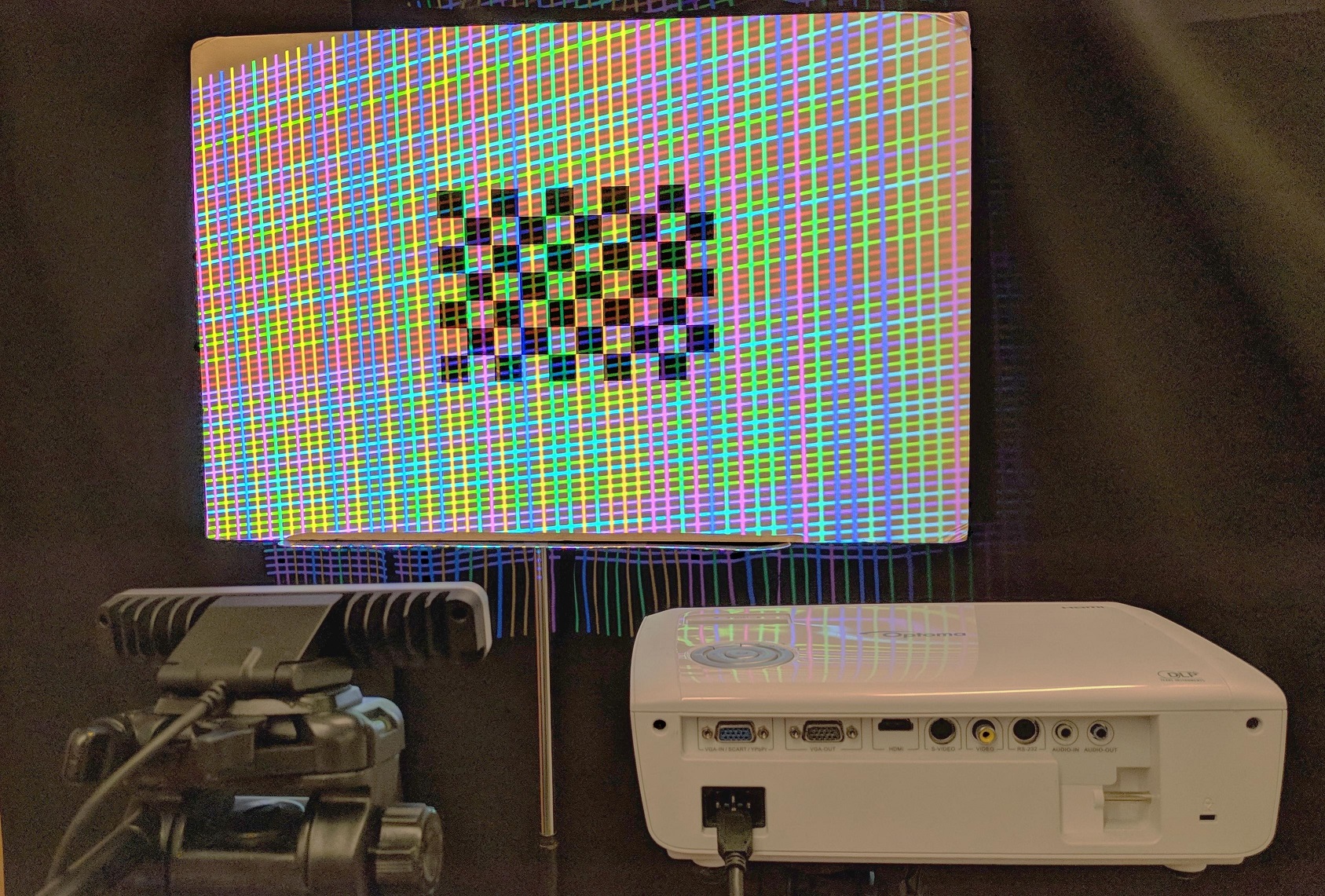}
		\caption[System setup]{System setup: a projector on the bottom-right, a camera on the bottom-left and calibration board with a checkerboard pattern attached to it.}
		\label{fig:setup}
		\vspace{-.5cm}
	\end{figure}

	A typical SL system consists of a calibrated camera and projector pair placed at a fixed distance and orientation as shown in \autoref{fig:setup}. Firstly, the projector projects known encoded patterns onto the target object, then the projected patterns are deformed according to the surface shape of the target object. Once the camera captures the deformed patterns, pixel correspondences between camera and projector can be established by matching the captured and projected patterns. In the end, the 3D coordinates of the deformed pattern pixels are triangulated, given the camera-projector parameters and pixel correspondences.
	
	Despite the simplicity, the 3D reconstruction precision of an SL system is highly dependent on the joint camera-projector pair calibration. Unlike a binocular stereo vision system, in an SL system, the projector is unable to capture images. So most SL calibration systems model the projector as an inverse camera that can ``see" the calibration target \cite{Moreno2012a, Zhang2006, Ben-Hamadou2013a}. Then it can be simply calibrated like a camera using Zhang's method \cite{Zhang2000a}.
	
	The above camera-projector framework requires correspondences between the projector image plane and a reference plane, which is usually approximated by a planar calibration board with a printed checkerboard, dots or circles pattern. The camera can be calibrated using these patterns. Afterwards, the projector projects encoded SL patterns onto the calibration board, and these patterns are then captured by the camera for calibrating the projector.
	
	
	In practice, the calibration board and printed pattern are hardly perfect planar due to manufacturing and/or glue. As pointed out by \cite{Zhang2000a, Sun2006, Albarelli2010, Strobl2011, Huang2013}, an imperfect calibration target may significantly impact the accuracy of Zhang's method. As most existing calibration methods rely on camera parameters to warp printed feature points to the projector image space, the camera error may be propagated to the projector image plane. This adds to the error of projector calibration that is again done by Zhang's method.
	
	To address this issue, we propose a novel additional step to jointly rectify the camera and projector models. Specifically, after calibrating the camera and projector using the traditional method, we put them into a bundle adjustment (BA) framework \cite{Furuakwa2009} for rectification, together with a scale regularization for further improvement.
	Another key component in our system is the reliable correspondence construction process. By using an efficient De Bruijn pattern \cite{Zhang2002, Huang2014} and a carefully designed keypoint extraction algorithm, our system provides reliable keypoint correspondence for the calibration algorithms. Moreover, being a single-shot per pose\footnote{Following \cite{Geng2011}, we call it \emph{single-shot} for conciseness in the rest of the paper.} solution, our system brings practical convenience over systems that require multiple shots of SL patterns for a single calibration board pose. This is particularly important for applications that require frequent re-calibrations, e.g., with the camera/projector moving around.
	
	To summarize, our system brings the following contributions:
	\begin{itemize}
		\item\vspace{-1.34mm} Our system explicitly deals with the noise in target planarity with a novel BA solution. This is the first such system for joint camera-projector calibration, to the best of our knowledge.
		\item\vspace{-1.34mm}  Unlike many existing methods, to calibrate the system, we apply points from an SL pattern rather than from a checkerboard. This strategy boosts both the number of feature points and their spatial distribution, and hence improves calibration robustness.
		\item\vspace{-1.34mm}  The proposed method performs camera-projector pair calibration with only a single-shot per pose, making it practically convenient in many applications. It can provide a flexible and accurate results even when the board is handheld.
	\end{itemize}
	
	The effectiveness of the proposed solution over existing ones is clearly demonstrated in our experiments on both synthetic and real data, especially when the calibration board is imperfect planar. In addition, the source code is publicly available at \url{https://github.com/BingyaoHuang/single-shot-pro-cam-calib}.
	
	
	In the rest of the paper, we summarize related work in \autoref{sec:related_work} and introduce our camera-projector calibration method in \autoref{sec:method}. Then, we report experiments in \autoref{sec:results}, and conclude this paper in \autoref{sec:conclusion}.

	\section{Related Work}\label{sec:related_work}
	
	Most existing camera-projector pair calibration methods apply Zhang's method \cite{Zhang2000a}, where the 3D-2D correspondences between the points on the calibration board and the projector image are computed by some transformations. Regardless of a  multi-shot or single-shot method, their transformations fall into one of the following methods: global homography \cite{Fiala2005a,Kimura2007, Zhang2007b,Ouellet2008, Drareni2009, Anwar2012, Orghidan2012, Huang2015a, Dhillon2015}, local homography \cite{Moreno2012a, Li2010}, direct pixel-to-pixel transformation \cite{Zhang2006} and incremental projector image pre-warp \cite{Audet2009a, Chen2013, Zhang2015, yang2016, Shahpaski_2017_CVPR}.
	
	It is worth noting that a global homography-based  method usually ignores both projector lens distortions and imperfect planarity of the calibration board. While the other three types of calibration methods can model projector lens distortion \cite{Zhang2006, Li2010, Moreno2012a, Chen2013, Zhang2015}, they are highly dependent on camera calibration accuracy. In addition, the imperfect planarity of calibration board is ignored in all the reviewed methods above, and such imperfectness can cause errors as pointed out in~\cite{Zhang2000a, Sun2006, Albarelli2010, Strobl2011, Huang2013} and illustrated in \autoref{sec:results}.
	
	Other than using Zhang's method, self-calibration algorithms \cite{Zhang2007b, Yamazaki2011, Li2017bb, willi2017} are capable of calibrating intrinsics and extrinsics of the camera-projector pair without a known planar target, instead a fundamental matrix or its variant is estimated using camera and projector image correspondences. With some priori of the intrinsics, \eg, unit aspect ratio and the principle point is assumed to be at the center of the image. However, these two assumptions are often violated, especially for projectors \cite{Moreno2012a}.
	
	Additional cameras can also be included to either reduce calibration board poses \cite{Bird2011} or allow arbitrary shapes as calibration targets \cite{Resch2015g}. However, these methods are even inflexible since they either require additional hardware or precise 3D measurement of a complex object, let alone multiple shots.
	
	\textbf{Multi-shot and single-shot methods}
	According to \cite{Geng2011}, SL-based camera-projector calibration methods can be categorized into two types: multi-shot \cite{Chen2013,Moreno2012a,Bird2011,Yamazaki2011,Zhang2006, Petkovic2016,yang2016, willi2017} and single-shot \cite{Zhang2002, Ben-Hamadou2013a,Anwar2012, Zhang2007b,Lee2007}. Specifically, multi-shot and single-shot indicate the number of SL pattern shots for \emph{each} calibration board pose, rather than the total number of shots for the whole process.

	Multi-shot methods project a sequence of patterns onto the calibration board, the patterns are encoded in Gray/binary code \cite{Moreno2012a, Petkovic2016, willi2017} or multiple phase shifting \cite{Zhang2006}, leading to a pixel-wise or even sub-pixel resolution. However, a disadvantage is that it is slow and computationally expensive due to multiple shots, \eg, \cite{Moreno2012a} requires about 20 shots and captures for each pose.
	Incremental methods \cite{Audet2009a, Li2010, Chen2013, Zhang2015, yang2016, Shahpaski_2017_CVPR} also belong to multi-shot, since the projected pattern is incrementally adjusted to fit the printed pattern until a perfect superimposition is achieved, which requires at least two shots per pose.
	

	Despite the correspondence accuracy, multi-shot calibration methods are both computationally expensive and memory inefficient compared with single-shot ones. Moreover, multi-shot methods are sensitive to motion; even a little shift or jitter between two consecutive captures can produce huge SL decoding errors due to pattern misalignment. For example, when a user holds the calibration board or a mobile camera-projector pair, it is very hard to ensure absolute stillness of the target between consecutive shots.
	
	Single-shot methods only require one shot per pose and adopt spatial multiplexed patterns including the De Bruijn sequence \cite{Salvi2004, Zhang2002, Huang2014,kawa08,Sagawa2009}, M-array \cite{Ben-Hamadou2013a,Zhang2007b}, checkerboard \cite{Anwar2012} and phase shifting fringes. The feature point correspondences are uniquely encoded in a single SL pattern. Consequently, single-shot SL allows for faster and more flexible camera-projector calibration than multi-shot SL.

	Compared with previous studies, our camera-projector calibration is simple and fast, and requires only one shot per calibration board pose. Moreover, it refines imperfectly calibrated camera and projector parameters due to imperfect planar calibration board using a bundle adjustment method. The experiments show that our method outperforms the other counterparts on both synthetic and real data.
	
	

	\begin{figure}[!t]
		\centering
		\vspace{-.5cm}
		\includegraphics[width=1.0\linewidth]{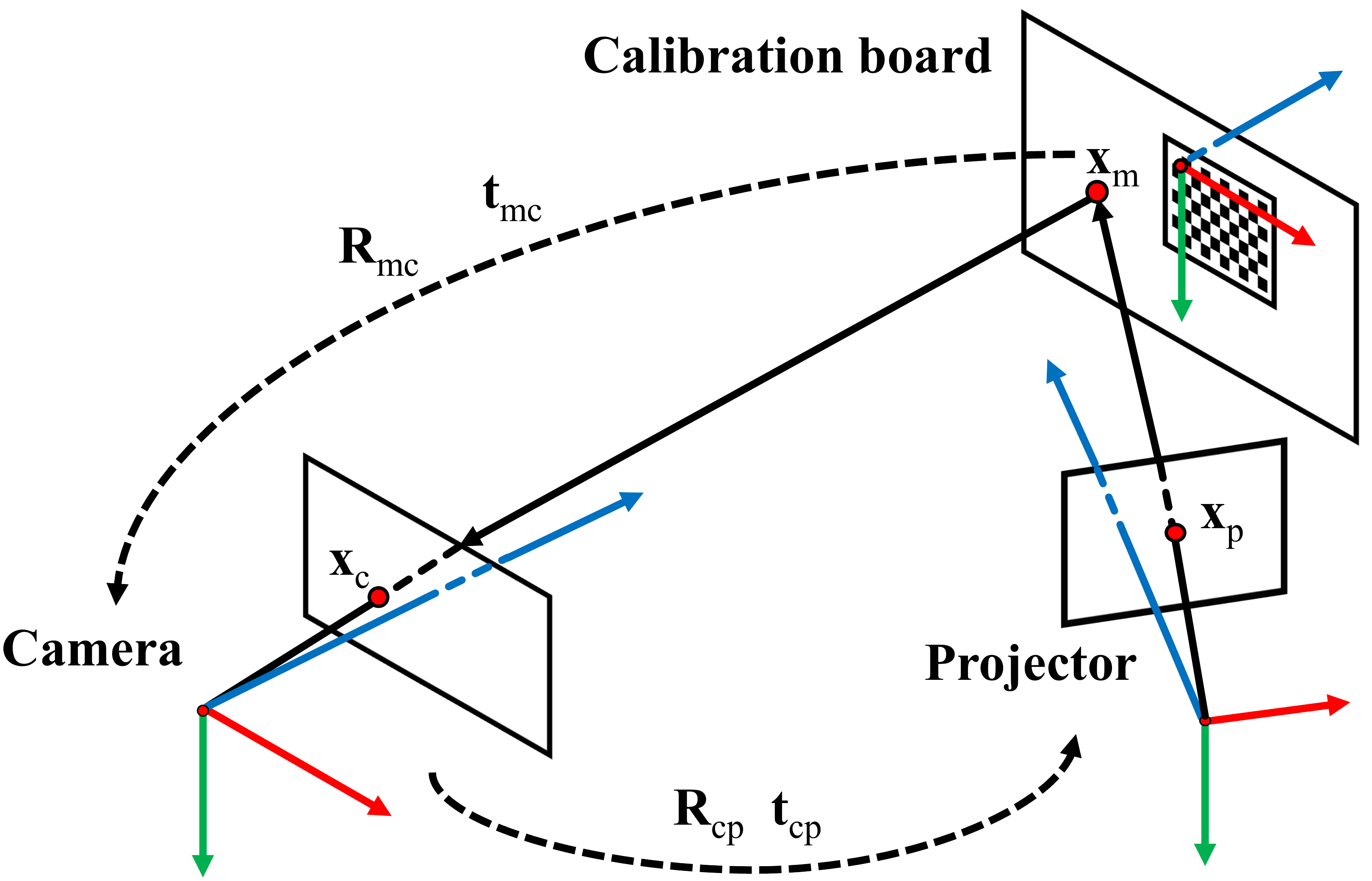}
		\caption{Coordinate system. The world origin is at the camera optical center. Red, green and blue axes represent X, Y and Z directions, respectively.}
		\label{fig:coordinate_system}
		\vspace{-.5cm}
	\end{figure}
	
	\vspace{-.1cm}
	\section{Method}\label{sec:method}
	\noindent\textbf{Notations.} Throughout the paper, we use the mathematical notations as shown in \autoref{tab:notation}. In addition we use subscripts $ _\mathrm{c} $, $ _\mathrm{p} $ and $ _\mathrm{m} $ for camera, projector and calibration board model space, respectively.	Thus, the subscript $ _{\mathrm{cp}} $ (or $ _{\mathrm{mp}}$) stands for the transformation from camera (or calibration board) coordinate system to projector coordinate system (\autoref{fig:coordinate_system}).
	
	\begin{table}[!h]
		\begin{center}
			\vspace{-.2cm}
			\caption{Notations}		\label{tab:notation}\vspace{-.2cm}
			\begin{tabular}
				{rcl}
				\hline
				\bf{Notation} & \bf{Example} & \bf{Meaning} \\
				\hline
				italic & $a,A,\dots$ & scalars \\
				\hline
				lower-case boldface & $\vec{a},\vec{b},\dots$ & vectors \\
				\hline
				boldface capital & $\mat{A},\mat{B},\dots$ & matrices \\
				\hline
				calligraphic  & $\mathcal{A}, \mathcal{B}, \dots$ & sets \\
				\hline
				index range & $ \mat{a}^{1:N} $ & $ \mat{a}^{1}, \mat{a}^{2}, \dots, \mat{a}^{N} $\\
				\hline
				dot & $ \dot{\vec{a}}, \dot{\vec{A}},\dots $ & initial guess\\
				\hline	
				hat & $ \hat{\vec{a}}, \hat{\vec{A}},\dots $ & estimation\\
				\hline	
				bar & $ \bar{\vec{a}}, \bar{\vec{A}},\dots $ & homogeneous coordinates\\
				\hline	
			\end{tabular}
		\end{center}
		\vspace{-.5cm}
	\end{table}
	
	\subsection{System Overview}
	
	\begin{figure*}[!t]
		\centering
		\vspace{-.8cm}
		\includegraphics[width=1.0\linewidth]{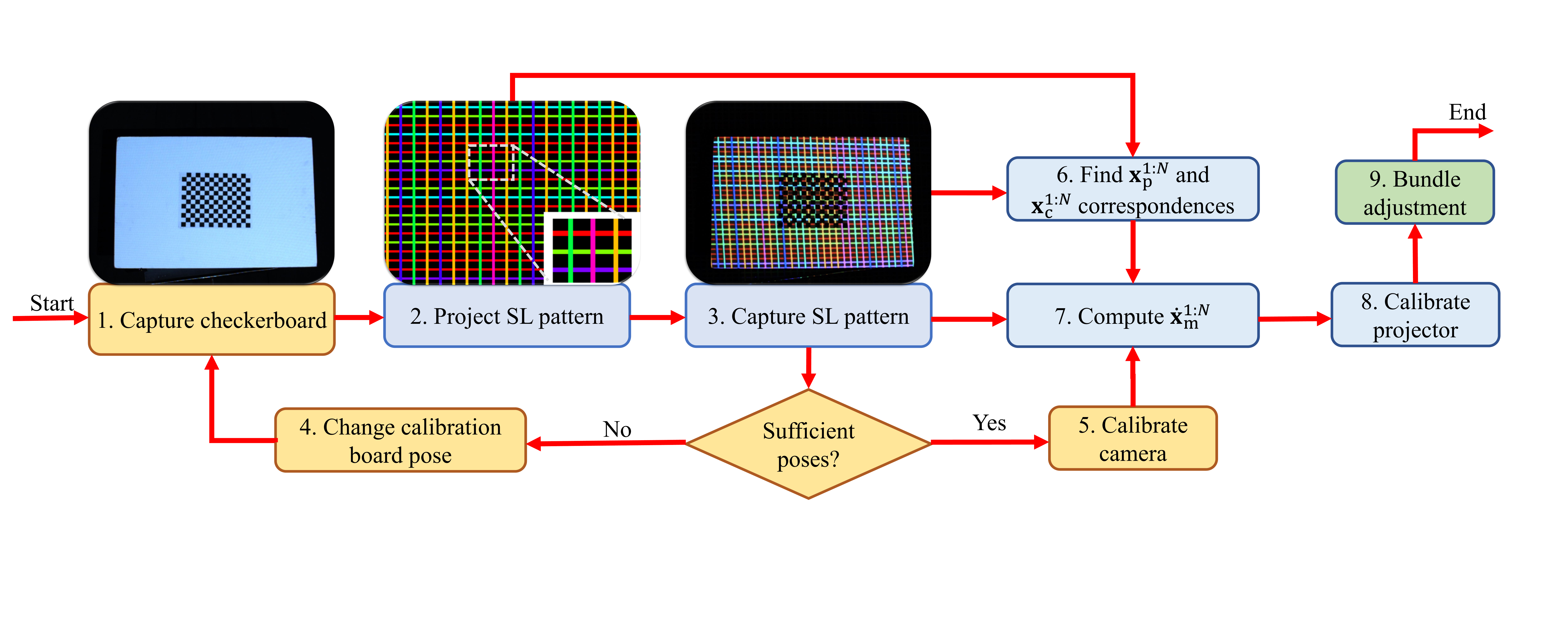}
		\caption[System flowchart]{System flowchart.  We divide the calibration algorithm into three procedures (\autoref{alg1}): \textbf{Yellow blocks}: camera initial calibration. \textbf{Blue blocks}: projector initial calibration and \textbf{Green block}: bundle adjustment. Best viewed in color.
		}
		\label{fig:systemoverview}
		\vspace{-.5cm}
	\end{figure*}

	Our camera-projector calibration system (\autoref{fig:setup}) consists of an RGB camera, a projector and a white board with a printed checkerboard pattern attached to it. As summarized in \autoref{alg1}, it contains three stages: (1) initial camera calibration using checkerboard images, (2) initial projector calibration using projected SL patterns, and (3) joint refinement of camera and projector parameters using bundle adjustment (BA).
	
	As shown in the system flowchart in \autoref{fig:systemoverview}, we start by capturing an image of the calibration board, then we project a color-encoded SL pattern to the calibration board and take an image of the superimposed SL pattern. We change the pose of the calibration board manually and repeat the steps above to get sufficient (at least three) pose samples. 
	Then, we first calibrate the camera using Zhang's method to get the initial camera model, including camera intrinsics and rotations and translations of each calibration board pose relative to the camera.

	We then undistort the captured SL images. After that, we decode the SL patterns in the camera image plane and find their correspondences to the original SL pattern in the projector image plane. Following that, we transform the SL points to the calibration board model space using rotations and translations obtained in last step. Note the SL points in the calibration board model space may also be erroneous due to inaccurate camera calibration.
	With these correspondences, we apply Zhang's method again to calibrate the projector. The relative rotation and translation between camera and projector are estimated using stereo calibration. Similar to camera parameters, the projector parameters obtained are also initial guesses and subject to propagated errors from camera calibration and imperfect planarity of the calibration board.

	Finally, we gather camera and projector parameters along with the SL points 
	to perform a BA refinement. This last step largely reduces errors in initial calibration (\autoref{subsec:ba}).

	\subsection{Structured Light Pattern}\label{sec:sl}
	
	To allow for single-shot calibration, we employ a spatial multiplexed SL technique and use only a single color-encoded pattern (\autoref{fig:systemoverview} steps 2-3). 
	The SL pattern is a variant of \cite{Salvi1998} composed of vertical and horizontal colored stripes with a De Bruijn sequence encoding. A De Bruijn sequence of order $n$ over an alphabet of $k$ color symbols is a cyclic sequence of length $k^{n}$ with a so-called window property that each subsequence of length $n$ appears exactly once \cite{db}.
	
	
	Let $\mathcal{C} = \{1, 2, ..., 8\}$ be the set of encoding color primitives, each number represents a different color. In particular, red (1), lime (3), cyan (5) and purple (7) are used for the horizontal stripes, while yellow (2), green (4), blue (6) and magenta (8) for vertical ones. In the inset of  \autoref{fig:systemoverview} step 2, the vertical color stripes are $(4,8,2)$ from left to right, the horizontal stripes are $(1,3,7)$ from top to bottom. 
	
	We employ De Bruijn encoding to both vertical and horizontal stripes, and construct a color grid with $m\times m$ intersections, where $m=k^{n}+2$, in our case $ k=4, n=3 $. More importantly, a unique $k$-color horizontal sequence overlain atop a unique $k$-color vertical sequence only occurs once in the grid. As shown in inset of \autoref{fig:systemoverview} step 2, this $ 3\times3 $ subset color grid appears only once in the whole pattern. We represent the color-coded pattern using an undirected graph	$\mathcal{G}=(\mathcal{V}, \mathcal{E})$, in which
	$\mathcal{V}=\{\vec{v}_{1}, \vec{v}_{2}, $ $\ldots$ $\vec{v}_{m\times m}\}$ is a set of graph nodes, which represent color stripes intersections, where		
	\begin{equation}
	\vec{v}_i =\{\vec{x}_\mathrm{c}(i), \vec{x}_\mathrm{p}(i), \vec{x}_\mathrm{m}(i) \},	
	\end{equation}	
	such that $\vec{x}_\mathrm{c}(i)= [u_\mathrm{c}(i), v_\mathrm{c}(i)]^T$, $\vec{x}_\mathrm{p}(i)=[u_\mathrm{p}(i), v_\mathrm{p}(i)]^T$, and $\vec{x}_\mathrm{m}(i)=[ x_\mathrm{m}(i), y_\mathrm{m}(i), z_\mathrm{m}(i) ]^T$ represent the coordinates of the $i^{th}$ node in, respectively, the camera image space, the projector image space and the calibration board model space.		
	$\mathcal{E} = \{\vec{e}_{00}, \vec{e}_{01}, \dots, \vec{e}_{ij}, \dots, \vec{e}_{m^2\times m^2}\} $ is the set of all edges representing color stripe segments. We have $\vec{e}_{ij} = \{\mathcal{L}, \tau\} $, where $\mathcal{L} $ is a list of pixels belonging to this edge and  $\tau\in\mathcal{C}\cup\{0\}$ is the color label of the edge (if the link exists) or 0 (otherwise).
	
	The correspondences between the camera captured image and the projected SL pattern is built by decoding the color codeword of the SL pattern. Since finding SL correspondences is not the main focus of this paper, we provide the details in the supplementary material.
	
	Once we have the camera and projector coordinates of all the nodes, we apply the homography $ \mat{H}^j $ to transform node points from the camera image plane to the calibration board model space, where $ \mat{H}^j $ is the transformation for the $j$-th pose estimated by initial calibration (\autoref{sec:init}).
	
	\begin{algorithm}[!t]		
		\caption{The proposed calibration algorithm}\label{alg1}		
		\baselineskip=12pt
		{			
			\begin{algorithmic}[1]				
				\State Input: camera captured images $ \mathcal{I}^{1:N}$
				\State Output: camera-projector pair parameters $ \vecg{\hat{\Psi}} $
				\State // {\it Stage 1. Initial Camera Calibration}
				\State  $ \mat{K}_\mathrm{c}, \vec{d}_\mathrm{c}, \mat{R}^{1:N}_\mathrm{mc}, \vec{t}^{1:N}_\mathrm{mc} \gets \text{ZhangCalib}(\mathcal{I}^{1:N})$
				
				\For{$j\gets 1$ to $N$}		
				\State \vspace*{-.7cm}
				\begin{align}\label{eq:homo} \mat{H}^{j} = \mat{K}_\mathrm{c} * [\vec{r1}^j_\mathrm{mc}, \vec{r2}^j_\mathrm{mc}, \vec{t}^j_\mathrm{mc}]& \quad\quad\quad\quad&\quad\quad\quad\quad
				\end{align}
				\vspace*{-.8cm}		
				\EndFor
				
				\State // {\it Stage 2. Initial Projector Calibration}				
				\For{$j\gets 1$ to $N$}
				\State $\vec{\bar{x}}^{j}_\mathrm{c} \gets \text{undistort}(\vec{x}^{j}_\mathrm{c}, \vec{d}_\mathrm{c})$
				\For{$i \gets 1$ to $M^j$}
				\State \vspace*{-.7cm} \begin{align}\label{eq:warp}\vec{\dot{x}}^{j}_\mathrm{m}(i) = \text{inv}(\mat{H}^{j})* \vec{\bar{x}}^{j}_\mathrm{c}(i)& \quad\quad\quad\quad\quad\quad\quad\quad
				\end{align}
				\vspace*{-.8cm}
				\EndFor
				\EndFor
				\State  $ \mat{K}_{\mathrm{p}}, \vec{d}_{\mathrm{p}}, \mat{R}^{1:N}_\mathrm{mp},
				\vec{t}^{1:N}_\mathrm{mp} \gets
				\text{ZhangCalib}(\vec{\dot{x}}^{1:N}_\mathrm{m}, \vec{\bar{x}}^{1:N}_\mathrm{p})$ \label{alg1:calibPrj}
				
				\vspace*{.1cm}
				\State	$ \vec{R}_{\mathrm{cp}} = \text{median}(\vec{R}^{1:N}_{\mathrm{mp}}*(\vec{R}^{1:N}_{\mathrm{mc}})^{T}) $ \label{alg1:initR}
				
				\vspace*{.1cm}
				\State	$ \vec{t}_{\mathrm{cp}} = \text{median}(\vec{t}^{1:N}_{\mathrm{mp}}-\vec{R}^{1:N}_{\mathrm{mp}}*(\vec{R}^{1:N}_{\mathrm{mc}})^{T}*\vec{t}^{1:N}_{\mathrm{mc}}) $ \label{alg1:initT}
				
				\vspace*{.1cm}
				\State // {\it Stage 3. Bundle Adjustment}
				\State $ \vecg{\dot{\Psi}} = [\mat{K}_{\mathrm{c}}, \vec{d}_{\mathrm{c}}, \mat{R}^{1:N}_\mathrm{mc}, \vec{t}^{1:N}_\mathrm{mc}, \mat{K}_{\mathrm{p}}, \vec{d}_{\mathrm{p}}, \mat{R}_{\mathrm{cp}}, \vec{t}_{\mathrm{cp}}, \vec{\dot{x}}^{1:N}_\mathrm{m}]$
				\State  $ \vecg{\hat{\Psi}} \gets \text{bundleAdjust}(\vecg{\dot{\Psi}})$\\
				\Return $ \vecg{\hat{\Psi}} $
			\end{algorithmic}		
		}
	\end{algorithm}

	\subsection{Initial Calibration}\label{sec:init}
	The camera and projector view spaces and calibration board model space follow a right hand coordinate system as shown in \autoref{fig:coordinate_system}. The world origin is at the camera optical center.
	
	\textbf{Camera and Projector Model}
	We employ the pin-hole model for both camera and projector calibration, with intrinsic matrices denoted by $ \mat{K}_\mathrm{c} $ and $ \mat{K}_\mathrm{p} $, respectively:
	\begin{equation}\label{eq:CamIntrinsics}
	\mat{K}_\mathrm{c} =
	\begin{bmatrix}
	f_{x}   & 0     & c_{x}\\
	0       & f_{y} & c_{y}\\
	0       & 0     & 1
	\end{bmatrix}
	, \quad
	\mathbf{K}_{\mathrm{p}} =
	\begin{bmatrix}
	f'_{x}   & 0     & c'_{x}\\
	0       & f'_{y} & c'_{y}\\
	0       & 0     & 1
	\end{bmatrix},
	\end{equation}
	where $ f_x, f'_x $ and $ f_y, f'_y $ represent camera and projector focal lengths in $ x $ and $ y $ directions. $ (c_x, c_y )$ and $(c'_x, c'_y )$ represent camera and projector image principle point coordinates. The camera and projector distortion coefficients are given by:
	\begin{equation}\label{eq:CamDist}
	\vec{d}_{\mathrm{c}} = [k_1, k_2, p_1, p_2], \quad \vec{d}_{\mathrm{p}} = [k'_1, k'_2, p'_1, p'_2],	
	\end{equation}
	where $ k_1, k'_1 $ and $ k_2, k'_2 $ are radial distortion factors;  $ p_1,p'_1 $ and  $ p_2,p'_2 $ are tangential distortion factors.
	In addition, we model extrinsics parameters, i.e., relative rotation  and translation  of the camera with respect to the projector as:
	\begin{equation}\label{eq:Prjr}
	\mathbf{r}_{\mathrm{cp}} = (r_x, r_y, r_z)^T, \quad \mathbf{t}_{\mathrm{cp}} = (t_x, t_y, t_z)^T.
	\end{equation}
	Note that $ \mathbf{r}_{\mathrm{cp}} \in \mathfrak{so}(3) $ is a rotation vector, i.e., the associated Lie algebra of rotation matrix $ \mat{R}_{\mathrm{cp}}  \in \mathrm{SO}(3)$.
	%

	\textbf{Camera Calibration}
	We first calibrate the camera using Zhang's method~\cite{Zhang2000a}, with all the checkerboard corner correspondences from camera images $\{\mathcal{I}^{1}, \mathcal{I}^{2}, \dots \mathcal{I}^{N}\}$ to the  calibration board model space. We obtain initial guess of camera intrinsics $ \mat{K}_\mathrm{c} $ and $ \vec{d}_\mathrm{c} $, as well as relative rotation $ \mat{R}^{j}_\mathrm{mc} $ and translation $ \vec{t}^j_\mathrm{mc} $  between the $ j^{th} $ calibration board pose and the camera view space. A homography $ \vec{H}^{j} $ between the calibration board and the camera image plane can then be calculated by \autoref{eq:homo}, where $ \vec{r1}^j_\mathrm{mc} $ and $\vec{r2}^j_\mathrm{mc} $ are the $ 1^{st} $ and $ 2^{nd} $ columns of $ \mat{R}^{j}_\mathrm{mc} $ of the $ j^{th} $ pose.

	\textbf{Projector Calibration}
	After initial camera calibration, we transform the SL pattern nodes from camera image space to calibration board model space by \autoref{eq:warp} in \autoref{alg1}, where 
	$ \vec{\bar{x}}^{j}_\mathrm{c}(i) $ is the undistorted homogeneous coordinates of node $ \vec{v}_i $ in the camera image space, imaged at the $ j^{th} $ pose. To be clear, we do not use checkerboard corners for projector calibration, instead we employ the SL nodes since they provide more robust and accurate initial guess.
	
	Once we obtained the node point pairs $ ( \vec{\dot{x}}_\mathrm{m}, \vec{\bar{x}}_\mathrm{p}) $, Zhang's method is applied to calibrate the projector parameters, as shown in line 15 of \autoref{alg1}. The relative translation and rotation between camera and projector are computed as shown in lines 16-17 of \autoref{alg1}.
	

	\subsection{Bundle Adjustment}\label{subsec:ba}	
	The imperfect planarity of the calibration board can bring errors to the initial calibration. Attacking this problem, given the initial camera and projector calibration, we propose a bundle adjustment (BA) algorithm (Stage 3 of \autoref{alg1}) on the initial intrinsics and extrinsics, as well as node point coordinates $ \vec{{x}}^{1:N}_\mathrm{m} $ subject to reprojection errors.
	
	Specifically, we set the world origin at camera optical center and let camera and projector parameters be:	
	\begin{align}
	\vecg{\Psi}_\mathrm{c}  &= (\mat{K}_{\mathrm{c}}, \vec{d}_{\mathrm{c}}, \vec{r}^{1:N}_{\mathrm{mc}}, \vec{t}^{1:N}_{\mathrm{mc}})\\
	\vecg{\Psi}_\mathrm{p}  &= (\mat{K}_{\mathrm{p}}, \vec{d}_{\mathrm{p}}, \mathbf{r}_{\mathrm{cp}}, \vec{t}_{\mathrm{cp}}),\label{eq:prjParam}
	\end{align}	
	where {$ \vec{r}^{1:N}_{\mathrm{mc}} $} and {$ \vec{t}^{1:N}_{\mathrm{mc}} $} are relative rotation and translation vectors of the calibration board with respect to the camera; $ \mathbf{r}_{\mathrm{cp}}, \mathbf{t}_{\mathrm{cp}} $ are relative rotation and translation of the camera with respect to the projector.
	
	Now, the camera-projector calibration problem can be formulated as minimizing the following BA cost:
	\begin{equation}\label{eq:objFunc}
	\{\vecg{\hat{\Psi}}_\mathrm{c}, \vecg{\hat{\Psi}}_\mathrm{p}, \vec{\hat{x}}^{1:N}_\mathrm{m} \}= \argmin_{\vecg{\Psi}_\mathrm{c}, \vecg{\Psi}_\mathrm{p}, \vec{x}^{1:N}_\mathrm{m}}\big(\text{cost}(\vecg{\Psi}_\mathrm{c}, \vecg{\Psi}_\mathrm{p}, \vec{{x}}^{1:N}_\mathrm{m})\big),
	\end{equation}
	More specifically, suppose the $j^{th}$ calibration board pose has $ M^j $ nodes imaged on the calibration board, and denote $n_p =\sum_{j=1}^{N}M^j$. The objective function is formulated as:
	\begin{equation}\label{eq:cost}
	\text{cost} = \sum_{j=1}^{N} \sum_{i=1}^{M^j}\big(\delta^{j}_\mathrm{c}(i) + \delta^{j}_\mathrm{p}(i) + \lambda\delta^{j}_\mathrm{m}(i) \big).
	\end{equation}
	The first two terms represent reprojection errors of the node $ \vec{v}_i $ in camera and projector image space:
	\begin{align}\label{eq:reprojErr}
	\delta^{j}_\mathrm{c}(i) &= \|\vec{x}^{j}_\mathrm{c}(i) - f(\vecg{\Psi}_\mathrm{c};\vec{x}^{j}_\mathrm{m}(i))\|^2\\
	\delta^{j}_\mathrm{p}(i) &= \|\vec{x}^{j}_\mathrm{p}(i) - f(\vecg{\Psi}_\mathrm{p}, \vec{r}^{1:N}_{\mathrm{mc}}, \vec{t}^{1:N}_{\mathrm{mc}};\vec{x}^{j}_\mathrm{m}(i))\|^2, \label{eq:deltaP}
	\end{align}
	where $ f:\mathbb{R}^3\mapsto \mathbb{R}^2 $ projects a node coordinate $ \vec{x}^{j}_\mathrm{m}(i) $ from the calibration board model space to the camera/projector image space using camera/projector parameters.
	
	In addition, we add a scale constraint that bounds the scale of model point coordinates during bundle adjustment:
	\begin{equation}\label{eq:constraint}
	\delta^{j}_\mathrm{m}(i)=\|\vec{\hat{x}}^{j}_\mathrm{m}(i) - \vec{\dot{x}}^{j}_\mathrm{m}(i)\|^2
	\end{equation}
	This term is necessary since the model point coordinates are coupled with extrinsic parameters $ \mathbf{r}_{\mathrm{cp}}$ and $ \mathbf{t}_{\mathrm{cp}} $. The original model point coordinates $ \vec{\dot{x}}^{j}_\mathrm{m}(i) $ are computed by \autoref{eq:warp}. We introduce a weight $\lambda $ to control the weight of the scale constraint, and empirically set $ \lambda = \exp(-\delta^{j}_\mathrm{m}(i)) $.

	We apply the trust-region-reflective algorithm to solve for \autoref{eq:objFunc}.  Since we introduce $ n_p $ node model coordinates $\vec{x}_\mathrm{m} $ to bundle adjustment, leading $ 3\times n_p $ extra  parameters to optimize, a sparse Jacobian matrix pattern is designed to speed up numerical finite derivative computation.

	\section{Experiments and Results}\label{sec:results}
	Our camera-projector pair consists of an Intel RealSense F200 RGB-D camera with image resolution of $ 640\times480 $, and an Optima 66HD DLP projector set to the resolution of $800 \times 600$, as shown in \autoref{fig:setup}. Note we only use RGB camera for calibration and reconstruction, the depth camera is employed only for generating ground truth.
	The distance between the camera and the projector is 1500 mm and all the calibration board poses are around 700 mm to 3000 mm in front of the camera-projector pair.
	
	\subsection{Evaluated Baselines}	
	We compare our method with three other methods: a generalized global homography method, a multi-shot local homography method \cite{Moreno2012a}, and a degenerated baseline of the proposed method.

	To compare with other camera-projector calibration methods (\eg, \cite{Audet2009a}), even if we replicate their calibration patterns (\eg, the ARTags pattern) and configurations, the obtained calibration points are different. It is hard to make a fair comparison this way. Instead, we generalize a method named \textbf{Global homography} to represent global homography-based methods in our experimental configuration and therefore we are able to use the same calibration points.
	For local homography-based method, we employ the popular system by \textbf{Moreno \& Taubin}~\cite{Moreno2012a} with default parameters. 

	In addition, we generate a degraded version of the proposed algorithm by excluding the BA stage, named as \textbf{Proposed w/o BA}. In particular, this baseline only includes $ \vec{r}^{1:N}_{\mathrm{mc}} $ and $ \vec{t}^{1:N}_{\mathrm{mc}} $ in the nonlinear optimization. Unlike \textbf{Global homography} that uses points from a checkerboard, the degraded method applies points from an SL pattern. This strategy boosts both the number of feature points and their spatial distribution, and hence is more robust and accurate.
	
	The three methods are tested together with the proposed one using both synthetic and real data. Root mean square (RMS) reprojection errors, 3D alignment errors, intrinsics and extrinsics errors are used as criteria. The synthetic data results (\autoref{fig:syn:general}-\autoref{fig:syn:prjIn}) and real data results (\autoref{fig:recon}) show clearly the benefits of the proposed method.
	

	\vspace{-.2cm}
	\subsection{Synthetic Data}\label{sec:synth_data}
	\begin{figure}[t]
		\centering
		\vspace{-.5cm}
		\includegraphics[width=1.0\linewidth]{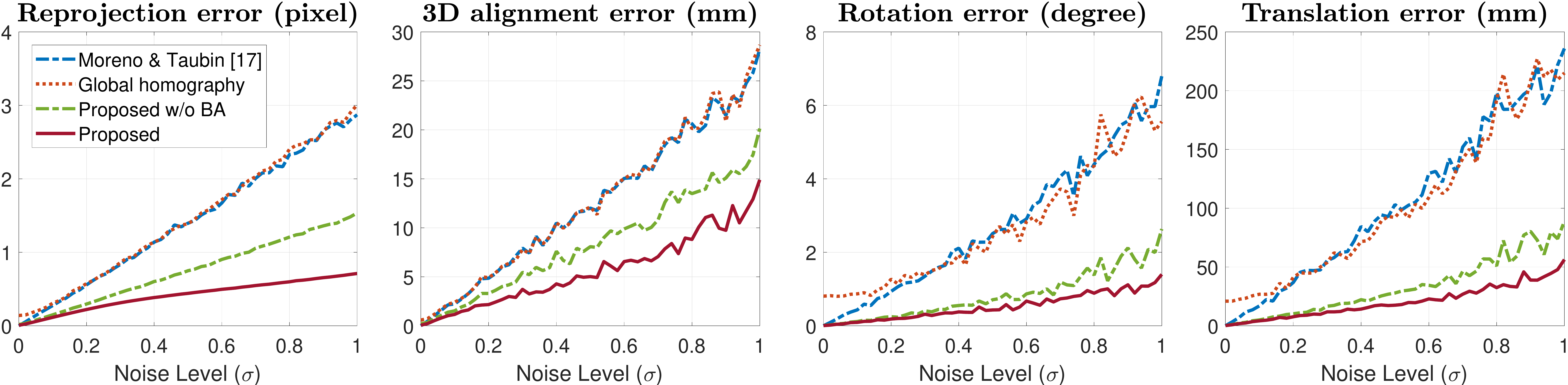}\vspace{-.2cm}		
		\caption{Synthetic data. Reprojection, 3D alignment, rotation and translation errors when noise level  $\sigma = 0 \rightarrow 1$.
		}
		\label{fig:syn:general}
		\vspace{-.2cm}
	\end{figure}
	
	\begin{figure}[t]
		\centering
		\includegraphics[width=1.0\linewidth]{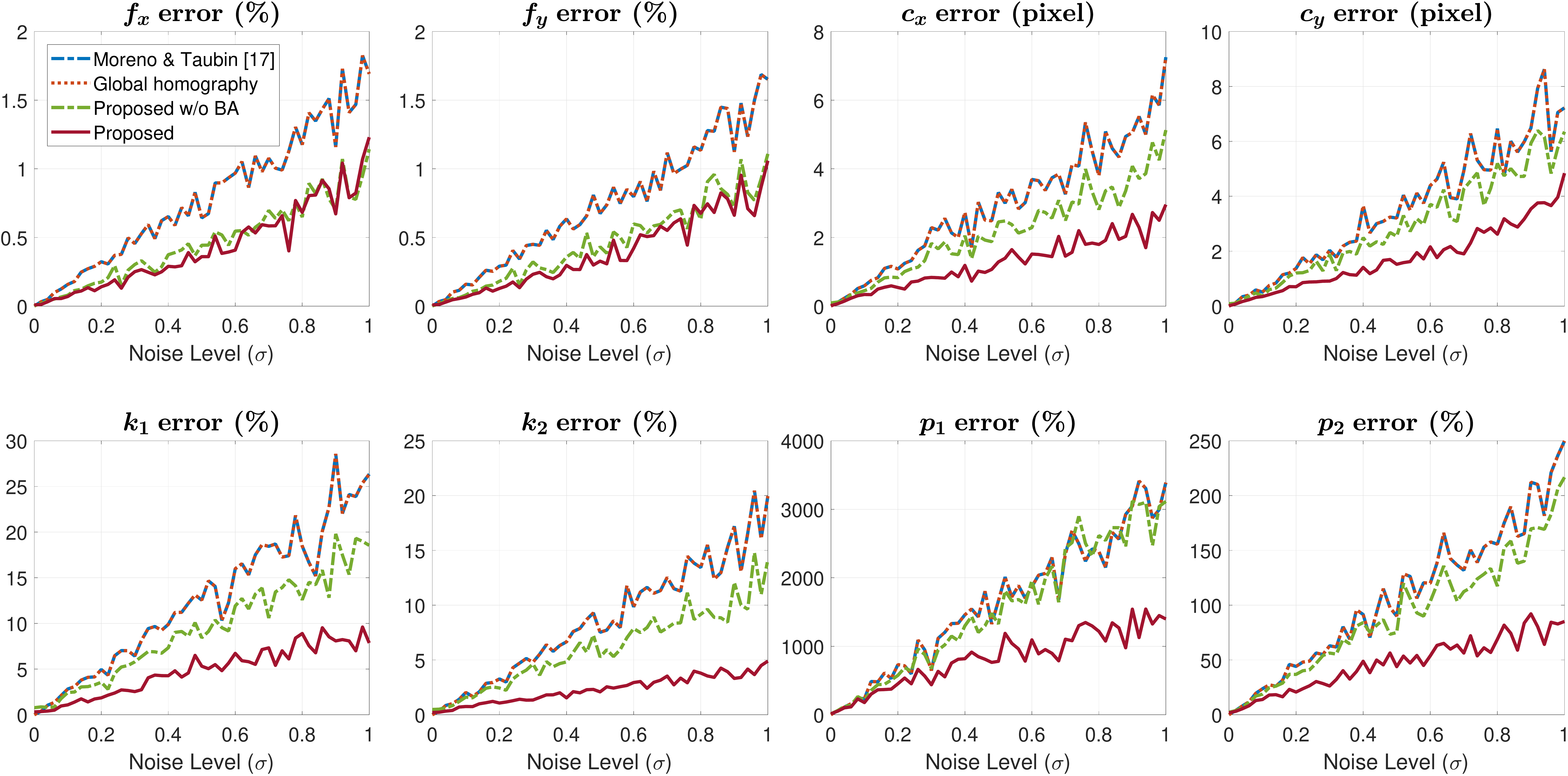}\vspace{-.2cm}	
		\caption{Synthetic data. Errors in \textbf{camera intrinsics} for different noise levels $\sigma = 0 \rightarrow 1$.
			$f_x$ and $f_y$ are camera focal lengths in two directions, $(c_x,c_y)$ is the camera image principle point, and 	$ k_1, k_2$ and $ p_1, p_2 $ are radial and tangential distortion factors,  respectively.
		}
		\label{fig:syn:camIn}
		\vspace{-.5cm}
	\end{figure}
	
	\begin{figure}[t]
		\centering
		\vspace{-.5cm}
		\includegraphics[width=1.0\linewidth]{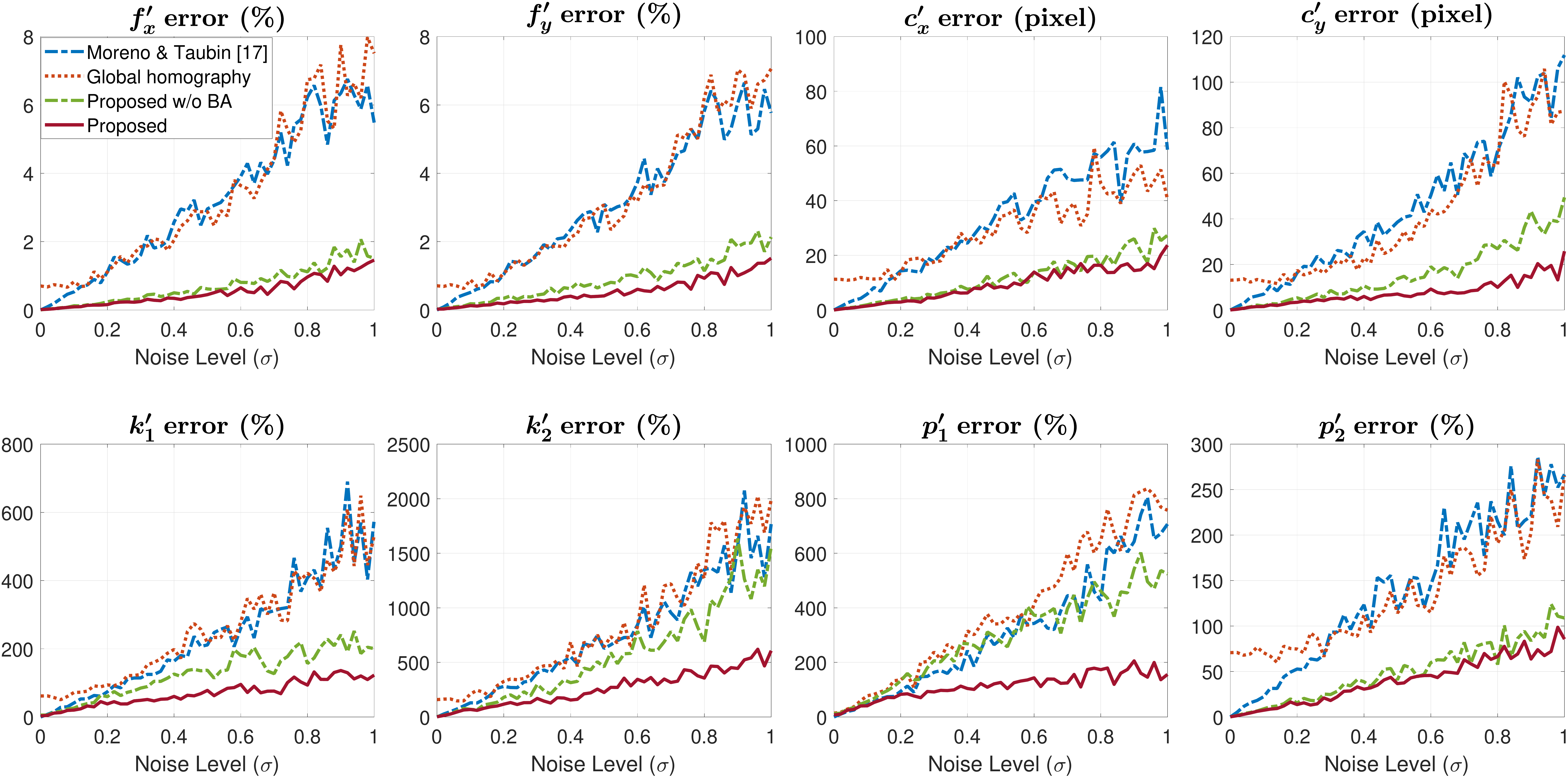}\vspace{-.2cm}	
		\caption{Synthetic data. Errors in \textbf{projector intrinsics} for different noise levels $\sigma = 0 \rightarrow 1$.
			$f'_x$ and $f'_y$ are projector focal lengths in two directions. $(c'_x,c'_y) $ is the projector image principle point, and $ k'_1, k'_2$ and $p'_1, p'_2 $ are radial and tangential distortion factors, respectively.
		}
		\label{fig:syn:prjIn}
		\vspace{-.2cm}
	\end{figure}
	
	To compare the proposed method with baseline methods statistically, we first use synthetic data as benchmarks, where the camera and projector intrinsics and extrinsics, checkerboard corners and calibration board geometry are known. In particular, synthetic data provides absolute ground truth and accurate error measurement.
	
	We start by generating the data by projecting the projector SL patterns to the world space. Each pair of node coordinate in projector image space $ \vec{x}_\mathrm{p} $ and projector optical center forms a ray that intersects with a set of predefined calibration boards, those intersections represent node's coordinates in the calibration board model space $ \vec{x}_\mathrm{m} $. Next we project $ \vec{x}_\mathrm{m} $ to the camera image space using pre-defined camera intrinsics and extrinsics, obtaining node's coordinates in camera image space $ \vec{x}_\mathrm{c} $. Finally, we follow the steps in \autoref{alg1} to calibrate our camera-projector pair.
	
	We add Gaussian white noise with zero mean and standard deviation of $\sigma$ to both camera and projector images. It is worth noting that, to simulate imperfect planarity, we also add Gaussian white noise to checkerboard and SL nodes in calibration board model space, whereas the noise units are in millimeters (mm). We generate the statistical benchmarks by inspecting the RMS reprojection errors, 3D alignment errors, intrinsics errors and extrinsics errors at each noise level $\sigma = 0 \rightarrow 1$. The 3D alignment errors are discrepancies between a synthetic 3D geometry and reconstructed 3D geometry.
	
	The experiments are performed 100 times for each noise level $\sigma$ and we plot the median of the errors as shown in \autoref{fig:syn:general} to~\autoref{fig:syn:prjIn}. The proposed method clearly outperforms the other three methods. Moreover, \textbf{Global homography}'s reprojection error, rotation error, translation error and some projector intrinsics errors are nonzero even when the noise level $ \sigma = 0 $ due to its inability to model projector lens distortions (\autoref{fig:syn:general}). In \autoref{fig:syn:camIn}, \textbf{Global homography}'s and \textbf{Moreno \& Taubin}'s curves overlap because they apply the same camera calibration method to the same set of checkerboard points.
	
	\subsection{Real Data}\label{sec:real_data}
	We evaluate our calibration using an imperfect planar white board with a printed checkerboard pattern glued to it.
	As shown in \autoref{tab:comparison} column 2, our method is able to refine imperfect planar points and inaccurate camera parameters using BA, thus leading to lower projector and stereo RMS reprojection errors than its counterparts. Note the stereo RMS reprojection error is the RMS of camera and projector reprojection errors.

	\begin{table*}[]
		\centering
		\vspace{-.8cm}
		\caption{Calibration RMS reprojection errors (pixels) (column 2) and reconstruction errors (mm) of real objects (columns 3-5).}\label{tab:comparison}
		\begin{tabular}{l|lll||lll|lll|lll}
			\hline
			\textbf{Method}                     & \multicolumn{3}{c||}{Reproj. errors (pixels)}                                                              & \multicolumn{3}{c|}{Paper box  (mm)}                                                                   & \multicolumn{3}{c|}{Plaster bust (mm)}                                                                & \multicolumn{3}{c}{Folded paper board (mm)}                                                          \\ \hline
			& Cam.                       & Pro.                           & Stereo                         & Mean                           & Median                         & Std.                           & Mean                           & Median                         & Std.                           & Mean                           & Median                         & Std.                           \\
			Moreno \& Taubin \cite{Moreno2012a} & \textbf{0.12}            & 1.59                        & 1.13                           & 8.47                           & 7.08                           & 5.93                           & 5.60                           & 4.72                           & 3.93                           & 9.82                           & 9.69                           & 5.72                           \\
			Global homography                                    & \textbf{0.12}            & 5.79                        & 4.09                           & 11.88                          & 11.94                          & 9.99                           & 9.81                           & 9.86                           & 4.85                           & 18.42                          & 19.91                          & 9.41                           \\
			Proposed w/o BA                                       & 0.42                     & 0.71                       & 0.58 & 6.78                           & 6.88                           & \textbf{4.10} & 6.10                           & 5.28                           & 4.16                           & 5.68                           & 4.86                           & 4.09                           \\
			Proposed                                             & 0.35                    & \textbf{0.64}               & \textbf{0.51}  & \textbf{5.60} & \textbf{4.59} & 4.70                           & \textbf{4.82} & \textbf{4.12} & \textbf{3.50} & \textbf{5.09} & \textbf{4.46} & \textbf{3.53} \\ \hline
		\end{tabular}
		\vspace{-.4cm}
	\end{table*}
	
	\begin{figure}[!t]
		\centering
		\includegraphics[width=1\linewidth]{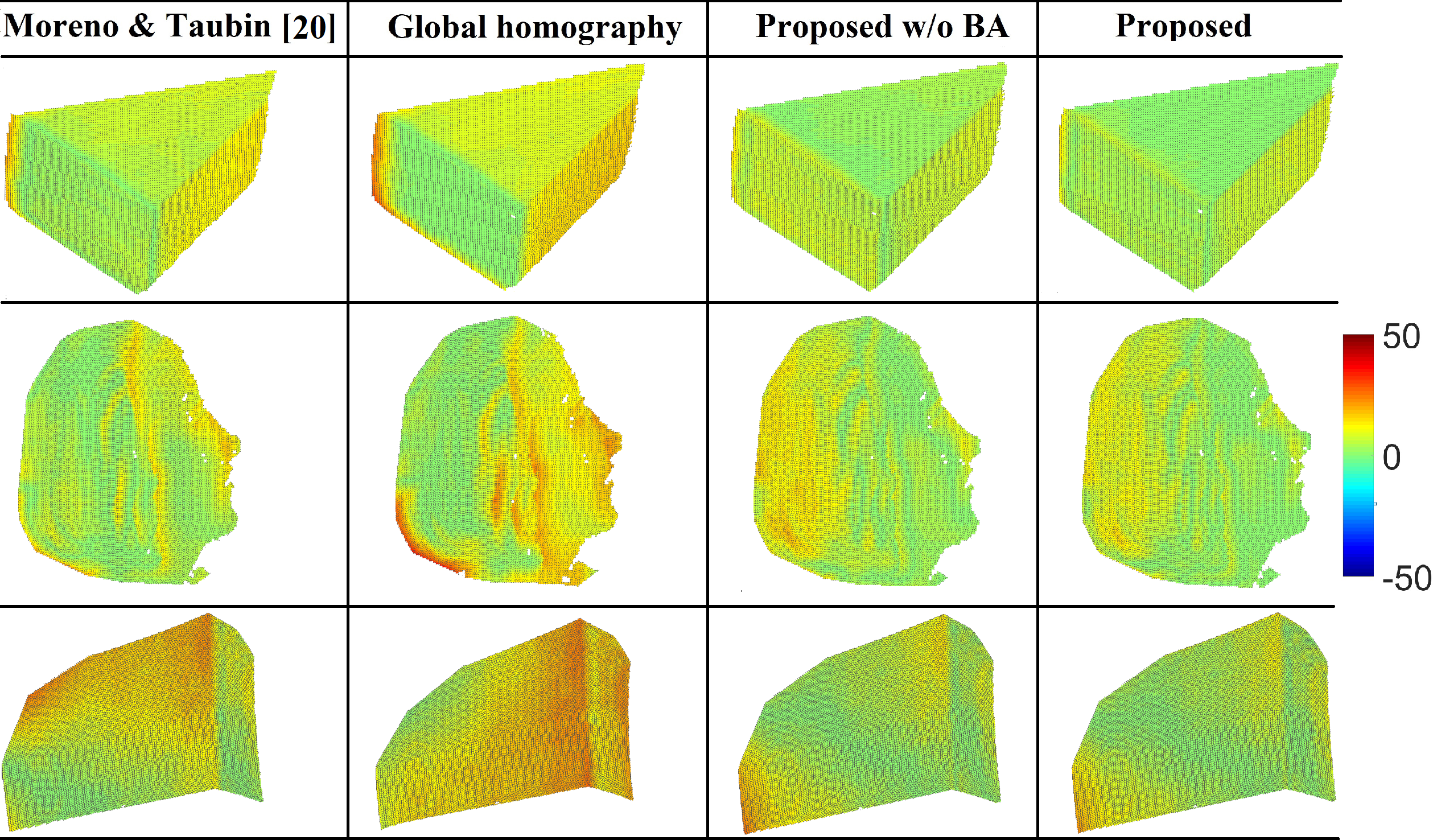}	
		\caption{ Real data. Reconstructed paper box ($ 1^{st} $ row),  plaster bust ($ 2^{nd} $ row) and folded paper board ($ 3^{rd} $ row) using a camera-projector pair calibrated by the four calibration methods. Reconstruction errors (mm) are shown in pseudocolor.}\label{fig:recon}
		\vspace{-.5cm}
	\end{figure}
	
	It is worth noting that the projector RMS reprojection errors of the first two methods are high for two reasons: (1) they use Zhang's method to calibrate the camera-projector pair and thus suffer from imperfectness in the planarity of the calibration board. (2) The errors of extrinsics, i.e., $ \mat{R}_{\mathrm{cp}} $ and $ \mat{t}_{\mathrm{cp}} $, propagate to the projector (see \autoref{eq:prjParam} and \autoref{eq:deltaP}), leading to enlarged high RMS reprojection errors.
	
	One may notice that our camera reprojection error is a bit higher than the other two methods. This is because SL nodes are used for camera-projector calibration, while the reprojection errors are based on nodes rather than checkerboard points. Namely, reprojection errors solely are not sufficient to represent calibration accuracy if different set of points are employed.  
	
	Thus, we evaluate  reconstruction errors by comparing the reconstructed point cloud with the ground truth. We first employ the calibration data from the four methods to reconstruct a point cloud using SL. Then the  reconstruction errors are calculated as the RMS discrepancies between the SL reconstructed point cloud and the RGB-D camera captured point cloud. As shown in \autoref{fig:recon}, a paper box, a plaster bust and a folded paper board are reconstructed using the calibration data of the four evaluated methods. The statistics of reconstruction errors are given in \autoref{tab:comparison}, columns 3-5, the \textbf{proposed w/o BA} method outperforms \textbf{Global homography} and \textbf{Moreno \& Taubin} \cite{Moreno2012a}, since it applies SL nodes to calibration. The \textbf{proposed} method outperforms the degraded version, proving that BA is able to compensate for imperfect nonplanarity.
	
	Our method outperforms both global and local homography-based methods on projector and stereo RMS reprojection errors and  reconstruction errors. In practice, our single-shot method also overcomes the drawbacks of requiring many shots per pose, whereas \textbf{Moreno \& Taubin} \cite{Moreno2012a} needs 20 shots per pose, and one local homography per checkerboard corner. Additionally, Audet \etal \cite{Audet2009a} and Yang \etal \cite{yang2016} need at least two shots per pose for prewarp and additional time for incremental adjustment.

	\vspace{-.2cm}
	\section{Conclusions}\label{sec:conclusion}
	In this paper we present a flexible single-shot camera-projector calibration method.	Compared with existing calibration systems, our method has two advantages: (1) Both synthetic and real data demonstrate that our method can refine imperfectly calibrated camera/projector parameters and imperfect planar calibration board points, thus leading to higher calibration accuracy and robustness against noises in planarity. (2) Requiring only a single shot of SL pattern per pose, our system enables fast and efficient calibration, especially in applications that need frequent re-calibration. Furthermore, the one-shot calibration provides a flexible and accurate results even when the board is handheld.
	
	

	\vspace{1mm}\noindent\textbf{Acknowledgement.} We thank the anonymous reviewers for valuable suggestions. Liao was supported in part by the China National Key Research and Development Plan (No. 2016YFB1001200).
	
	\vspace{-.1cm}
	\small{
		\bibliographystyle{abbrv-no-doi}
		\bibliography{ref}
	}

\end{document}